\pdfoutput=1

\documentclass[11pt]{article}

\usepackage[usenames,dvipsnames]{color}
\usepackage[final]{acl}
\usepackage{times}
\usepackage{latexsym}
\usepackage{makecell}

\usepackage{tabularx}
\usepackage{makecell}
\usepackage{multirow}
\usepackage{booktabs}
\usepackage{graphicx} 
\usepackage{circuitikz} 
\usepackage[T1]{fontenc}

\usepackage[utf8]{inputenc}

\usepackage{microtype}

\usepackage{inconsolata}

\usepackage{graphicx}
\usepackage{tabularx}
\usepackage{makecell}
\usepackage{amsfonts}
\usepackage{graphicx,tabularx,ragged2e,booktabs}
\title{Large Vision-Language Models for Knowledge-Grounded Data Annotation of Memes 
}

\author{Shiling Deng \\
  University of Copenhagen \\
  \texttt{shilingdeng7187@gmail.com} \\\And
  Serge Belongie \\
  University of Copenhagen \\
  \texttt{s.belongie@di.ku.dk} \\\And
  Peter Ebert Christensen \\
  University of Copenhagen \\
  \texttt{pec@di.ku.dk} \\}

\begin{document}
\maketitle
\begin{abstract}
Memes have emerged as a powerful form of communication, integrating visual and textual elements to convey humor, satire, and cultural messages. Existing research has focused primarily on aspects such as emotion classification, meme generation, propagation, interpretation, figurative language, and sociolinguistics, but has often overlooked deeper meme comprehension and meme-text retrieval. To address these gaps, this study introduces ClassicMemes-50-templates (CM50), a large-scale dataset consisting of over 33,000 memes, centered around 50 popular meme templates. We also present an automated knowledge-grounded annotation pipeline leveraging large vision-language models to produce high-quality image captions, meme captions, and literary device labels overcoming the labor intensive demands of manual annotation. Additionally, we propose a meme-text retrieval CLIP model (mtrCLIP) that utilizes cross-modal embedding to enhance meme analysis, significantly improving retrieval performance. Our contributions include:(1) a novel dataset for large-scale meme study, (2) a scalable meme annotation framework, and (3) a fine-tuned CLIP for meme-text retrieval, all aimed at advancing the understanding and analysis of memes at scale. Source code can be found in our GitHub repository\footnote{\url{https://github.com/Seefreem/meme_text_retrieval_p1}}. 
\end{abstract}

\section{Introduction}



In recent years, memes have become one of the most impactful forms of communication on social media, playing a crucial role in how people express their opinions, emotions, and cultural commentary (\citealp{vyalla2020memeify}; \citealp{pramanick2022multimodal}). Since the rise of social media platforms in the late 2000s, memes have evolved into a pervasive digital phenomenon, with millions of users creating, sharing, and interpreting them daily (\citealp{hwang2023memecap}; \citealp{buchel2012internet}; \citealp{tanaka2022learning}). 
Due to their widespread use, understanding the types of memes shared by online communities provides valuable insight into the cultural and social dynamics of these groups \citep{gal2016gets}. 
Thus understanding the cultural context in various modalities makes the interpretation of memes a fascinating yet challenging task for both humans and machines.
For instance, a meme might take an image from a well-known TV series and pair it with new text to produce a humorous twist on a familiar situation (\citealp{sharma2020semeval}; \citealp{sharma2023you}). The incongruity between the image and text, such as in popular meme formats featuring specific characters like ``One Does Not Simply'' from The Lord of the Rings or ``Not sure If'' from Futurama (see Table \ref{tab:types} (a)) often drives the humor. 
Other types of memes differ from typical images by their reliance on context and figurative language. For instance, in Table \ref{tab:types}(c), a literal image caption might describe a meme as ``colored trashcans,'' whereas a meme caption, the meaning of the meme, might convey a deeper meaning, such as ``comparing teammates to garbage cans, implying they are poor players despite having fancy character skins.'' Understanding a meme requires knowledge of figurative language (\citealp{liu2022figmemes}; \citealp{chakrabarty2021mermaid}; \citealp{chakrabarty2021figurative}), cultural references, and the interplay between image and text. This complexity underscores the need for models that can effectively handle both the visual and textual aspects of memes.\\

\begin{table*}[ !ht]
\small
\centering

\begin{tabularx}{\textwidth}{|>{\centering\arraybackslash}X| >{\centering\arraybackslash}X| >{\centering\arraybackslash}X|>{\centering\arraybackslash}X|}
 
 \hline
 Meme Type & \textbf{(a)} Character Macro & \textbf{(b)} Format Macro & \textbf{(c)} Memetic Images \\
 \hline

  Meme Examples&  \includegraphics[width = \hsize]{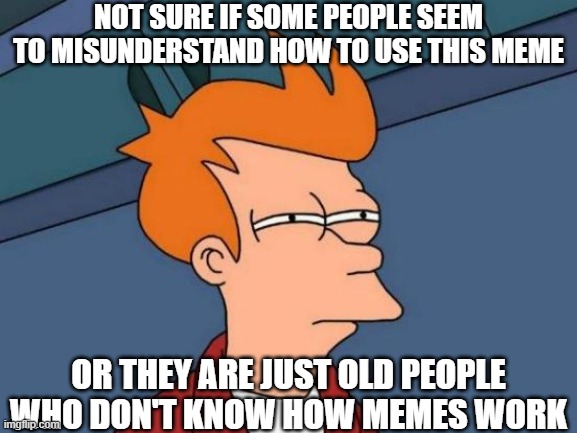} &  \includegraphics[width = 0.6\hsize]{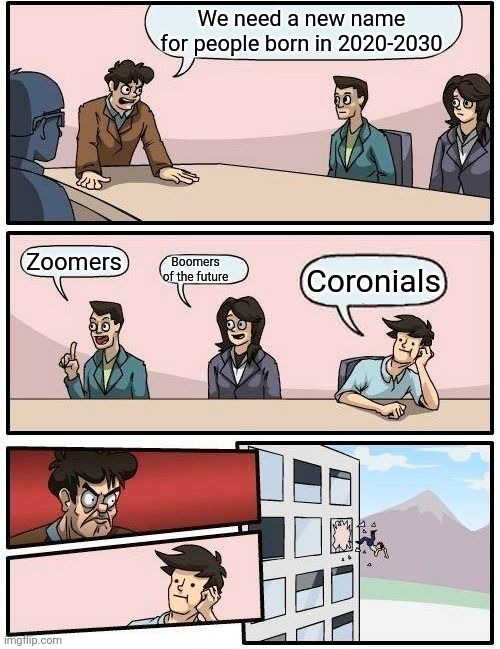} &  \includegraphics[width = \hsize]{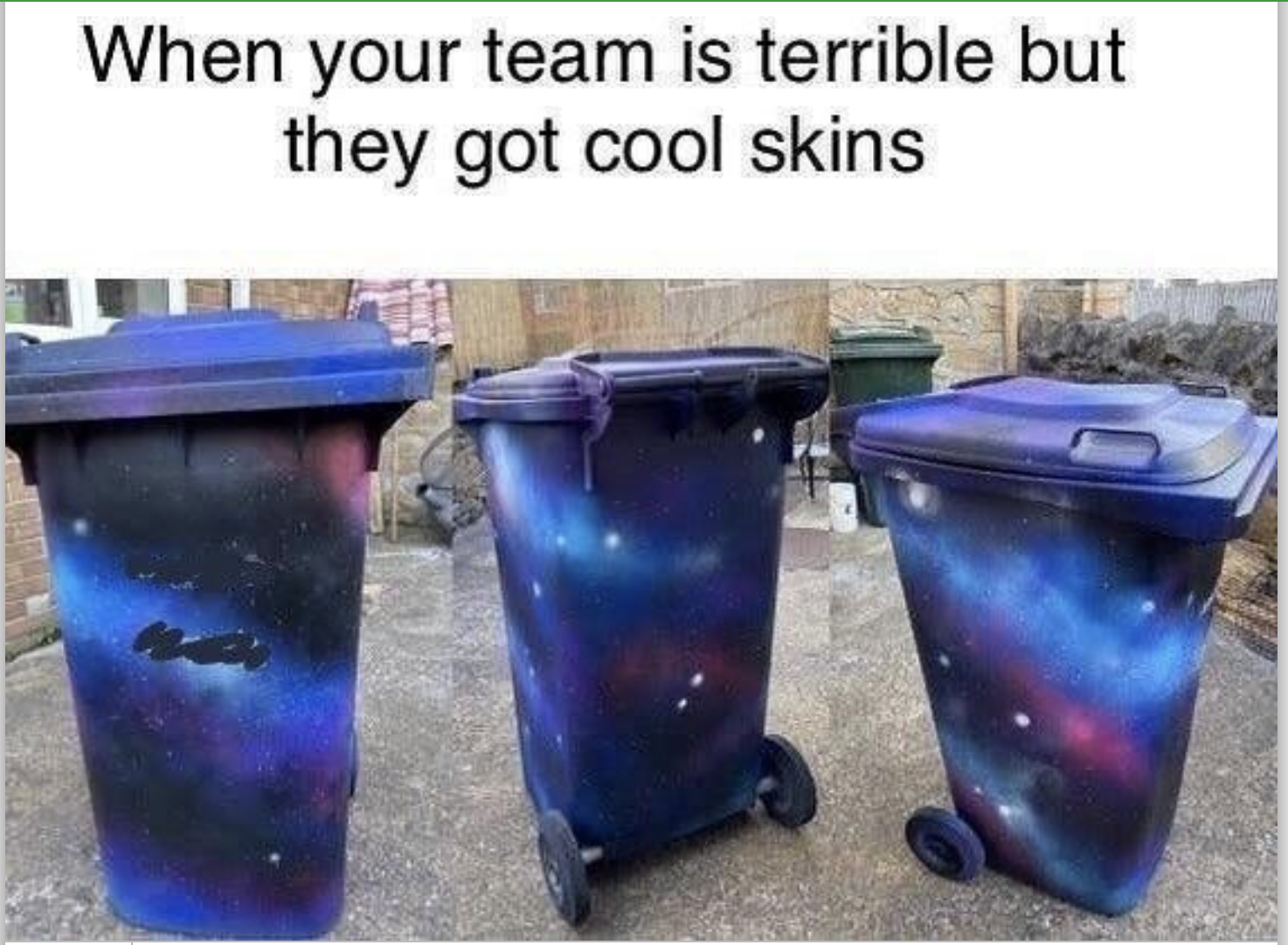} \\
\hline
\makecell{Memetic Element \\ (Location) \\ \textit{Novel Element}} & Futurama Fry (Background) \textit{Text captions} & Boardroom Meeting  (Background) \textit{Text Caption} &  Coloured Trashcans (Image)  \textit{Text clusters}\\
\hline
\makecell{Type is annotated\\by our method} & Is annotated &  Is annotated& Is not annotated \\
\hline

\end{tabularx}
\caption{Examples of selected meme types from the meme typology found in \cite{hazman2024makesmemememeidentifying} that are currently capable of being annotated using our method.
Each meme shown here is labeled with the \textit{memetic element} (and its location) each shares with, and the \textit{novel elements} that distinguish it from its related memes or meme templates. 
Note that other meme types that we do not consider in this study can be found, such as  ``Meme Trends'' and ``Superimposed images'' in \cite{hazman2024makesmemememeidentifying}. 
\vspace{-5mm}
\label{tab:types}}\end{table*}

Existing research in meme analysis has often relied on complicated pipelines involving multiple steps, such as image encoding, text encoding, optical character recognition (OCR), and clustering, such as in (\citealp{bates2023template}). Notable examples include the use of models like CLIP \citep{radford2021learning} to generate embeddings that capture both image and text features, as well as pipelines that use perceptual hashing and fine-tuned models like RoBERTa \citep{liu2019roberta} and CLIP for semantic embedding (\citealp{zhou2023social}; \citealp{liu2022figmemes}). These approaches, while effective, are cumbersome and require several components to work, making the process complex and computationally expensive. This has led to a growing need for a simplified approach, such as a specialized meme embedding model capable of directly capturing both the visual and textual content of a meme in a unified manner. 

\begin{figure}[t]
  \includegraphics[width=\columnwidth]{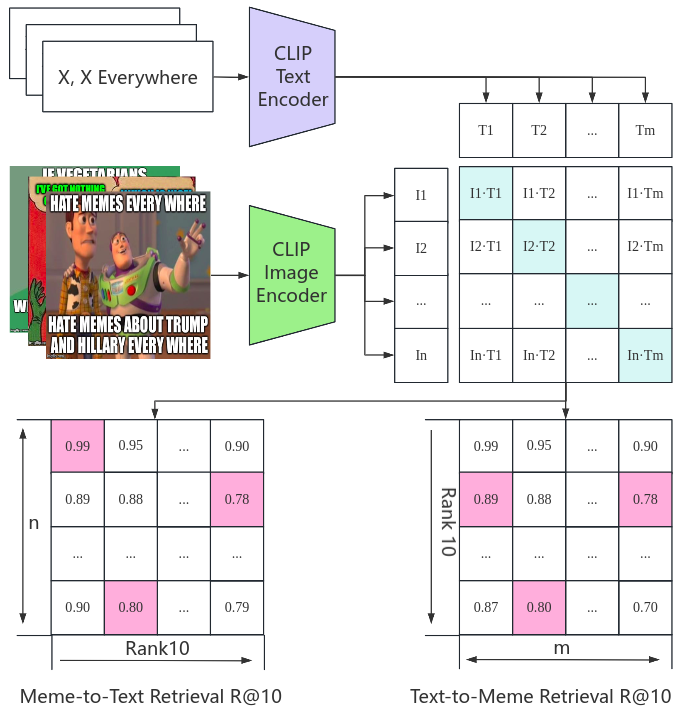}
  \caption{Diagram of meme-text retrieval, illustrated with Recall@10 for clarity, where $n$ and $m$ represent the numbers of memes and texts respectively. Meme and text embeddings are normalized to compute cosine similarities. The light blue cells indicate true matches in the similarity matrix, while the pink cells highlight true matches within the ranked lists.}
  \label{fig:meme-text-retrieval-schematic}
  \vspace{-5mm}
\end{figure}

Recent advancements in deep learning, particularly in multimodal learning, have led to the development of powerful models like CLIP and ALIGN\citep{jia2021scaling}, which align image and text representations in a shared latent space \citep{cao2022image}. These models have achieved state-of-the-art results in image-text retrieval benchmarks, but their performance in meme-specific contexts leaves room for improvement. Our pilot study is illustrated in Figure \ref{fig:meme-text-retrieval-schematic}, albeit at Recall@10 for illustration purposes, and results are reported in Table \ref{tab:pilot_study} which shows that CLIP outperforms ALIGN in meme-caption-to-meme and meme-to-meme-caption retrieval on MemeCap\citep{hwang2023memecap}, there is still a significant space for improvement. We found that both models are able to match text to image objects and embedded text, the text on an image, but still fall short of capturing the full semantic richness of memes. 
One step has been the recently proposed meme typology and meme identification protocol that aim to help interpret meme types and in the creation of meme datasets \cite{hazman2024makesmemememeidentifying}. They found that among the most popular 7 meme classification datasets more than half (50.4\% ) of the samples were not related to memes, highlighting the need for a large-scale dataset which fully contains memes. \\

\begin{table}
    \vspace{-5mm}
  \centering
  \begin{tabular}{l p{1cm} p{1cm} } 
    \hline
    {Model} & {T2M R@1}  & {M2T R@1}\\ 
    \hline
    ALIGN-base\footnotemark[1] & 0.539  & 0.568 \\ 
    Fine-tuned ALIGN  & 0.512 & 0.490 \\ 
    CLIP-ViT-B/32\footnotemark[2]  & 0.457  & 0.518  \\ 
    CLIP-ViT-L/14@336px \footnotemark[3] &0.648 &0.712\\ 
    \hline
    Fine-tuned CLIP &\textbf{0.760} &\textbf{0.780}\\ 
    \hline
  \end{tabular}
  \caption{Pilot study on MemeCap showing the R@1 score of both meme-to-text retrieval and text-to-meme retrieval, on a scale of $0-1$. Text refers to meme captions, and the meme is an image. 
  }
  \label{tab:pilot_study}
    \vspace{-5mm}

\end{table}
\footnotetext[1]{\url{https://huggingface.co/kakaobrain/align-base}}
\footnotetext[2]{\url{https://huggingface.co/openai/clip-vit-base-patch32}}
\footnotetext[3]{\url{https://github.com/openai/CLIP/blob/main/clip/clip.py}}

To address the challenge, we emphasize the need for a new and large annotated dataset. Existing meme datasets can be broadly categorized based on their focus areas, such as sentiment analysis, emotion detection, captioning, and sociolinguistics. However, these datasets are either limited in scope, or contain few examples of the same meme template. As memes evolve rapidly, there is an urgent need for a comprehensive dataset that captures the diversity and nuances of memes. Furthermore, most current labelled meme datasets are insufficient for training large models, as they often lack the scale, diversity or does not have a large variety of memes following the same template required to generalize effectively to new contexts.

To this end, we introduce a new dataset, Classic-Memes-50-templates (CM50), which consists of 33,172 memes collected from ImgFlip\footnotetext[4]{\url{https://imgflip.com/}}, focusing on instances of 50 popular meme templates. This dataset is designed to provide a rich resource for training and evaluating models on meme understanding tasks. Given the impracticality of manually annotating such a large dataset, we developed an automated annotation pipeline using GPT-4o-2024-08-06\footnotetext[5]{\url{https://openai.com/api/pricing/}}, by grounding our model in only 50 expert annotations for meme templates, which has shown promising results in generating accurate image captions, meme captions, and text extractions, albeit with some limitations in understanding literary devices.
The primary contributions of this paper are threefold:
\begin{itemize}   
\item [1)]The introduction of a new dataset, CM50, comprising 50 classic meme templates; 
\item [2)]The development of a framework for the automatic labeling of new templatic memes (memes that are templates or the instances of meme templates), facilitating scalable data collection;
\item [3)]The creation of a meme-text retrieval model that aims to improve the efficiency of large-scale meme analysis.
\end{itemize}

\section{Related Work}

\subsection{Meme Captioning and Generation}
The paper MemeCap \citep{hwang2023memecap} introduces a novel task: meme captioning, which aims to understand and interpret memes by examining their visual and textual components. They present a dataset called Memecap, containing 6.3K memes with annotations such as image captions, meme captions, titles, and metaphors. These annotations enable exploration of the complexities of visual metaphors and the interplay between text and images in memes. The study reveals that current vision-language models struggle with meme captioning, likely due to their lack of deep contextual knowledge, and our research builds upon this work to enhance model performance.
Dank Learning\citep{tolunay2018dank} presents a method for meme generation using CNNs and LSTMs to create humorous embedded text into meme templates, utilizing a dataset of 400,000 image-label-embedded text triplets. 
MemeCraft\citep{wang2024memecraft} further advances the field by using large language models (LLMs) and visual language models (VLMs) to generate context-driven memes that advocate social causes, employing user prompts and visual descriptions. The paper highlights safeguards against harmful content, showcasing the potential of LLMs for creative and socially impactful meme generation.

\subsection{Meme Propagation and moderation}
Memes that spread across various social media platforms might propagate differently between groups or platforms \cite{MemeKGGroundingDiscord}. This was investigated using 29160 meme posts from Discord and Reddit by grounding these in their knowledge graph whose data, akin to \cite{bates2023template}, is from KYM, in addition to ImgFlip, Twitter and Reddit.
News stories which contain certain narratives \cite{christensen20222} have also been shown to be capable of exhibiting meme like behavior in that the best narratives are replicated and spread while others disappears \cite{harlow2013facebook}.
Memes that are successful at spreading can have powerful effects when it resonates with people through storytelling, as the story can contain certain narratives which can cause reactions and changes in peoples behaviors \cite{harlow2013facebook}
Additionally malicious groups can seek to spread harmful or otherwise toxic content, which has prompted researching into new methods and datasets to intervene such behavior such as MemeGuard \cite{jha-etal-2024-memeguard}. They provide a dataset of 5.8K memes to benchmark toxic memes and their interventions, as well as their MemeGaurd framework that produces an intervention given a meme. One component in Memegaurd (VLMeme) aims to generate contextual information about the meme by training on the meme captions from MemeCap
\cite{hwang2023memecap}. 
Our work provides a way to scale up such efforts to large datasets of memes which share the same template but contain different content as we automate generation of meme captions, to help such model gain a deeper understanding of memes and their specific meanings.


\subsection{Sociolinguistics Study}
Memes can show sociolinguistic traits like focusing on particular geographical events, ethnicity, how people speak and in-group communication 
(\citealp{memeling}, \citealp{Holm_2021_sociolinguistic_memes}).
One such place to where in-group communication and discussion on events happens is on Reddit \cite{christensen2022}.
One paper \citep{zhou2023social} examines how memes reflect sociolinguistic variation by analyzing their multimodal structure of images and text. Using the SEMANTICMEMES dataset of 3.8 million Reddit memes, the authors created a pipeline involving clustering of meme instances into templates and semantic variables. They used fine-tuned RoBERTa\citep{liu2019roberta} and CLIP\citep{radford2021learning} models to extract visual and textual features, creating semantic clusters to understand social language patterns across communities. While the paper provides a way to cluster similar memes into templates they do not provide a way to use them for labeling memes.



\subsection{Meme Interpretation and discovery}
The KYMKB\citep{bates2023template} and KERMIT\citep{grasso2024kermit} papers highlight the value of integrating external knowledge into meme classification tasks. KYMKB introduces a comprehensive knowledge base of memes and meme templates, including information about their origins and examples, to enhance meme interpretation and labeling through the Template-Label-Counter model. This approach emphasizes the significance of contextual information in understanding memes. KERMIT focuses on harmful meme detection by injecting harmfulness-related knowledge into a multimodal classification model, thereby improving the accuracy of detecting harmful memes. These works demonstrate that incorporating additional contextual knowledge is crucial for effective meme interpretation and classification. Inspired by these insights we utilizing meme template knowledge from KnowYourMeme in our research.


\subsection{Figurative Language Understanding}
The field of figurative language understanding in memes has been advanced by V-FLUTE\citep{saakyan2024v} and FigMemes\citep{liu2022figmemes}, which explore machine learning models' ability to handle figurative language in visual and multimodal content. V-FLUTE focuses on visual figurative language entailment with a dataset of over 6,000 instances, using a human-AI collaboration framework to generate textual explanations and evaluate models like Llava\citep{liu2024improved}. FigMemes, on the other hand, presents a dataset of over 5,000 politically-opinionated memes annotated with six common figurative language types, providing comprehensive evaluations of unimodal and multimodal models. 
Our work builds on these efforts by using Large Visual-Language Models for data annotation and generating literary device labels to enhance understanding of figurative elements in memes.

\section{CM50: 50 Classic memes}

\begin{figure}[h]
    \vspace{-2mm}
    \centering
    \includegraphics[width=\columnwidth]{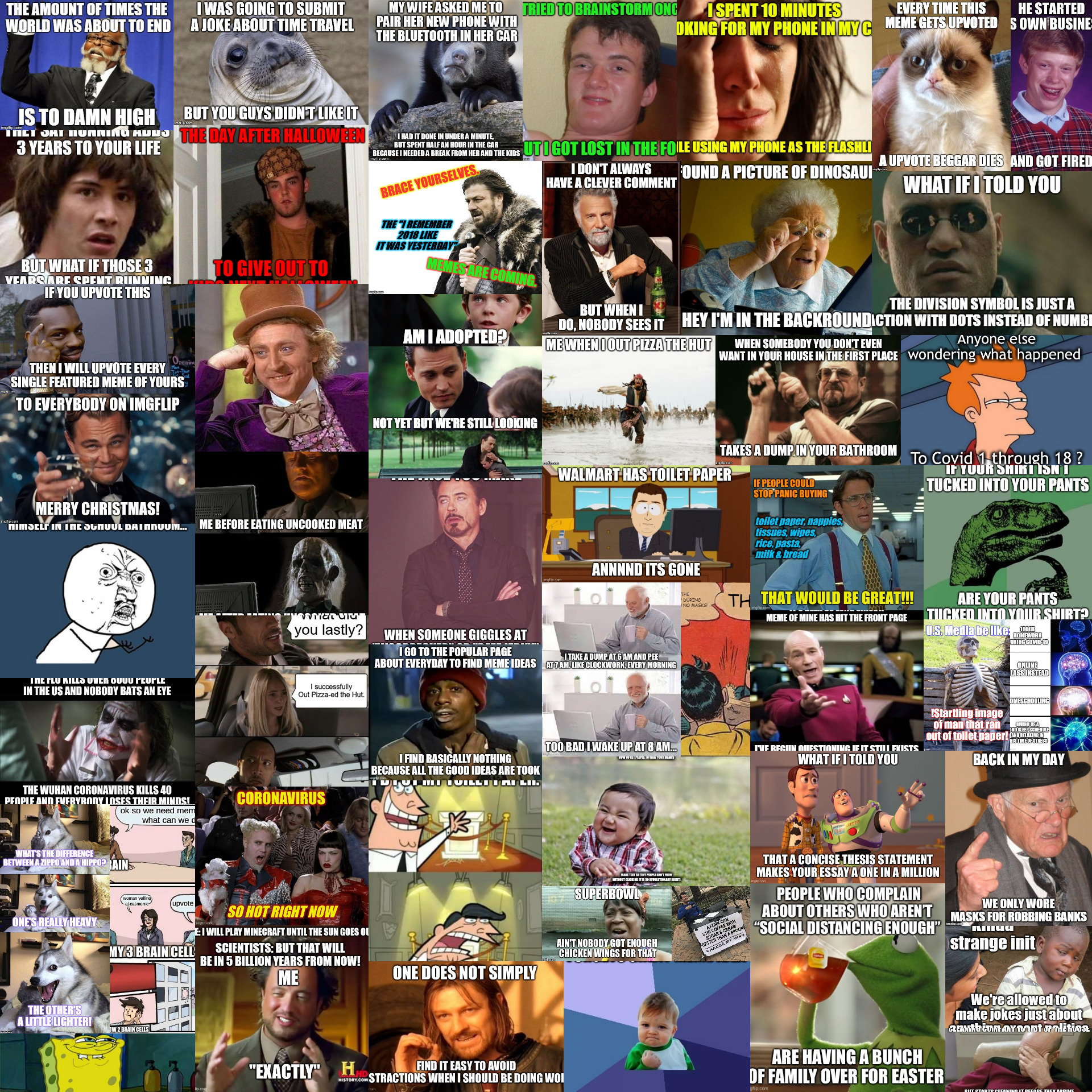}
    \vspace{-2mm}
    \caption{Collage of memes found in CM50}
    \label{fig:meme_collage}
\vspace{-5mm}
\end{figure}

\subsection{Dataset Details}
To conduct our study on memes we focus our attention to ImgFlip, one of the largest sources of memes which have been used by others such as \cite{decodingmemes2024} and  IMKG \cite{Tommasini2023IMKG}.
We notice that while (\citealp{hwang2023memecap}, \citealp{tanaka2022learning}, \citealp{wang2024memecraft}) have mentioned an available snapshot dataset from ImgFlip (e.g. the ImgFlip575K dataset.)\footnote{\url{https://github.com/schesa/ImgFlip575K_Dataset}} only \cite{defactify2023Yu} have used to conduct research on it, by using it for ensamble learning for the memotion 3 task. 

\begin{figure*}[t]%
    \centering
    \includegraphics[width=160mm]{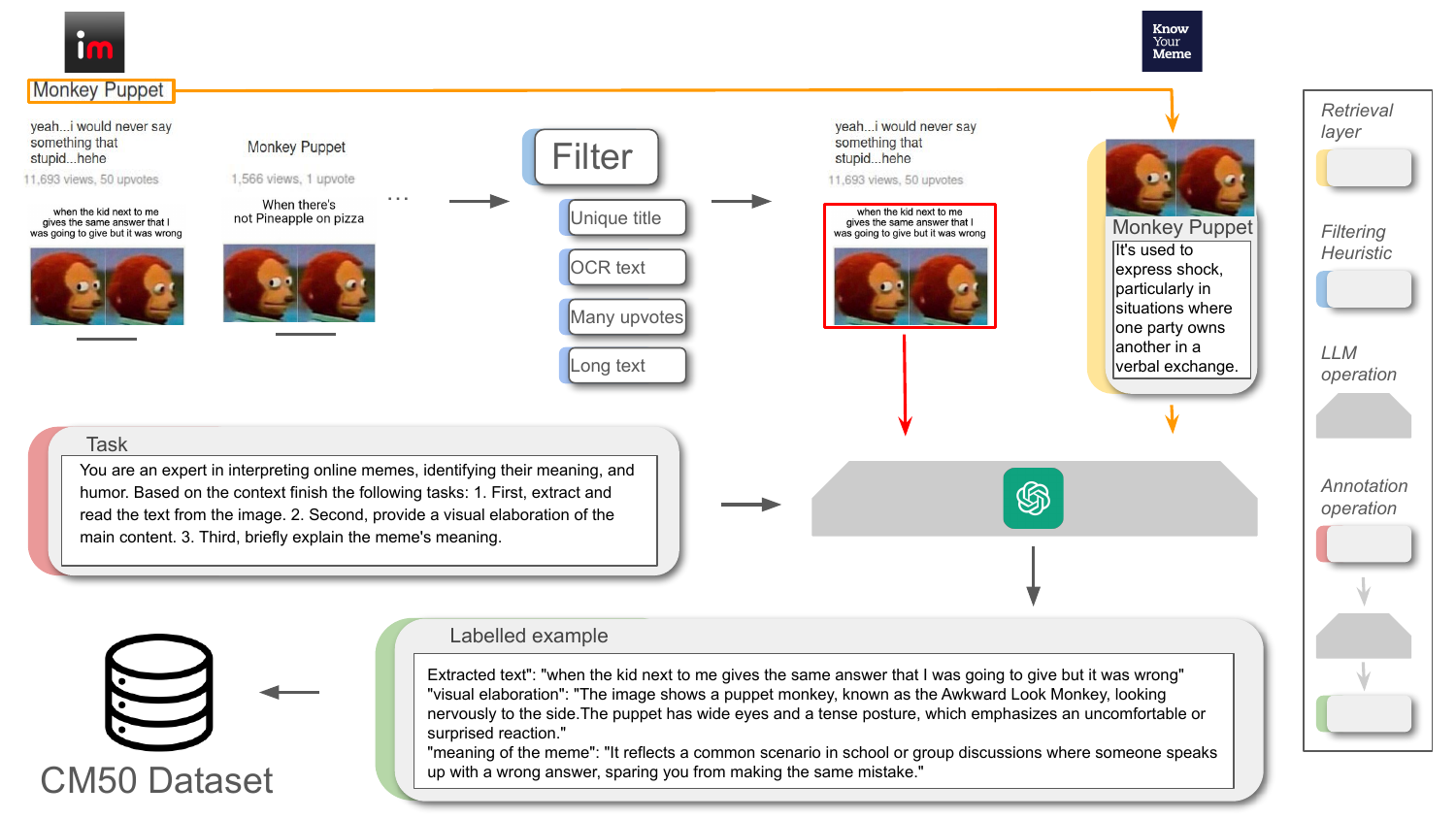}
    \caption{Example data annotation using our pipeline}
    \label{fig:data_pipeline}
        \vspace{-3mm}
\end{figure*}

One major issue is that the dataset is unfiltered and can contain misused or unpopular memes (i.e memes with a number of downvotes or views respectively), so we filter the dataset to contain at least 150 memes with text of sufficient length and a title that is different than the base template and constrain ourselves to the top 50 templates,  see Figure \ref{fig:data_pipeline} for details.
As such we end up with 33,173 memes, which are split into 31,823 examples for training and 1,350 for validation, and get their template metadata from KYMDB. Contextual details (the \textit{About} section of templates) were added to prompts for data annotation to generate the image caption, meme caption and literary devices. 
Each data item has the template of the meme, title of the meme, image caption, meme caption, embedded text and literary devices. In our dataset, we extended the literary devices to 26, of which 12 which can be mapped to the original 6 labels as found in prior work \cite{liu2022figmemes}, as we found the 6 categories limiting to describe the memes in our dataset. The annotation statistics are shown in Figure \ref{fig:annotation_statistics}. While some annotations exceed the maximum input length limit of CLIP, most satisfy the limitation. The literary device labels are highly imbalanced, partly due to the varying number of instances across different templates. For example, the large volume of instances for the ``But That's None of My Business'' template (as shown in Figure \ref{fig:data_imbalance}) may significantly inflate the number of irony labels, same as ``First World Problems'' to exaggeration labels. The distribution of memes across the 50 templates provided in Appendix \ref{sec:CM50_Dataset_details}. 
Instances of meme templates are shown in Figure \ref{fig:meme_collage}.



\subsection{Determining Annotation Tools}
Studies have shown that ChatGPT-3.5 and GPT-4 are widely used for data annotation and are more accurate than humans in some tasks (\citealp{he2024if}; \citealp{tan2024large}). GPT-4o, as one of the most advanced visual-language models (VLMs), is capable of complex understanding, including humor. This makes it an excellent candidate for automatic annotation. Notably, in \citep{saakyan2024v}, researchers leveraged GPT-4 to explain memes, suggesting that GPT-4o would demonstrate similar capabilities. However, the cost associated with using GPT models is a drawback. Therefore, we evaluated the open-source LLaVA-1.6 \citep{liu2024improved} \footnote{\url{https://github.com/haotian-liu/LLaVA?tab=readme-ov-file}}, a visual-language multimodal model, as a cost-effective alternative.
To find the optimal model for data annotation, we experimented with prompt engineering on GPT-4o and LlaVA-1.6, aiming to develop effective prompts for meme interpretation.
\subsubsection{Prompt Development}


Inspired by approaches in Prompt Engineering Guide\footnote{\url{https://www.promptingguide.ai/}}, we ask the model, GPT-4o, to explain the meme first, and then, to finish the five specific tasks: image captioning, extracting embedded text, meme captioning, literary device labeling, and emotion labeling. 
We found that by simply asking the model to interpret the meme, it recalled related information about the meme. This initial explanation served as additional context, enhancing the model's performance on the subsequent five tasks. We employed a similar iterative approach for refining prompts for LlaVA-1.6. Final Model Prompts are available in Appendix \ref{sec:Prompts_For_Model_Selection}.
\subsubsection{Human Evaluation and Results}
We evaluated the models using a survey of five diverse memes. Human evaluators, unaware of the model source, selected the one or multiple best answers.
Key findings include:
\begin{itemize}    
\item [a)] GPT-4o: Favored by evaluators, particularly with template context. It effectively captured the humor, literary devices, and emotions but relied heavily on both visual content and embedded text.
\item [b)] LlaVA-1.6: Achieved 89.8\% of GPT-4o's performance with template context but dropped to 82.3\% without it. LlaVA showed issues with following complex instructions and a bias toward labeling the literary device as ``humor.''
\end{itemize}

GPT-4o proved to be the most robust model for meme annotation due to its accuracy, consistency, and alignment in handling multiple tasks. Thus, we selected GPT-4o as the final model for data annotation. Additionally, The template context is informative for smaller VLMs and can help the models better align with human preferences.\\
For more details about the survey and its findings, please refer to Appendix \ref{sec:Human_Preference_Evaluation}.

\subsection{Automatic Evaluation of Our Automatic Annotation Pipelines}
\subsubsection{Evaluation Method}
We began by evaluating our annotation method on both the Figmemes and MemeCap test sets, consisting of 1542 and 559 examples respectively. Figmemes uses only six literary devices (i.e., figurative language), so we modified the task prompts to include these specific labels, along with their definitions to improve clarity.
To further assess GPT-4o's annotation quality on our dataset, we selected meme instances of our templates from both test sets and evaluated our prompts on these examples. Due to a lack of suitable data for verifying emotion labels, we focused our evaluation on image captions, meme captions, embedded text extraction, and literary device labeling.
To identify templatic memes, we developed a two-stage pipeline. We adopted the TemplateLabelCounter (TLC) pipeline from KYMDB \citep{bates2023template}, which matches templates and instances. We formulated the matching as a retrieval task: the query is the meme instance, and the retrieved image is the template. However, TLC had low recall (R\@1) on MemeCap, so we incorporated LPIPS from \citep{zhang2018unreasonable}, a learned perceptual image patch similarity metric. LPIPS computes the perceptual loss between image patches. 
We used two methods from \cite{bates2023template}, the ``concatenated embedding'' and ``fancy fusion embedding'' using CLIP with a threshold of 30 and 1 respectively to ensure instances matched wit a template meme. Afterward we employed the LPIPS similarity  using a threshold of 1 and manually verified the results, with any mismatched pairs being removed. This produced 2 lists of template-instance pairs which we merged, resulting in 46 template-instance pairs in Figmeme and 42 in MemeCap.
Using these templatic memes, we employed the GPT-4o prompt to annotate them. We then calculated automatic metrics including BLEURT, BERTscore (F1-macro), ChrF, ROUGE-L, and BLEU-4, based on the annotated features.

\subsubsection{Results}
\paragraph{Figmemes}
As shown in Table \ref{tab:Prompt_Engineering_on_Figmemes_Test_Set_Evaluation}, the prompt designed for the six literary devices achieved a macro F1-score of 0.39. This result is lower than the original paper's score, suggesting that GPT-4o performed worse compared to models trained on Figmemes' training set. To refine the prompt for literary device labeling, we selected 11 representative memes from the dataset and optimized our prompt. The initial prompt achieved a macro F1-score of 0.40 on these memes.
The best performance, as seen in Table \ref{tab:Prompt_Engineering_on_Figmemes_Test_Set_Evaluation}, was obtained with a prompt that included definitions of the literary devices. For few-shot prompting, three meme examples were selected based on their label amounts, label types, and common mistakes made by the model (e.g., misunderstanding comparisons). Interestingly, GPT-4o did not learn significantly from these examples, as the macro F1-score did not improve. Upon close analysis, we found that GPT-4o tends to over-interpret memes and often tries to identify as many figurative elements as possible. Since few-shot learning proved ineffective, domain adaptation might be necessary, although that is beyond the scope of this paper. We also observed that providing definitions for literary devices alongside other tasks in the prompt led to confusion, so the prompt used here excludes captioning tasks.
As shown in Table \ref{tab:Prompt_Engineering_on_Figmemes_Templatic_Memes}, we further evaluated the best prompt on templatic memes to assess the impact of meme template context. We also tested our own label set on these memes, mapping them to the six basic labels Figmemes for a better comparison. Despite the relative macro F1 score, the template context is shown to be informative in literary device labeling. Refer to Appendix \ref{sec:Prompt_Engineering_on_Figmemes_Templatic_Memes} for examples of the prompts and result details.
\paragraph{MemeCap}
As shown in Table \ref{tab:Prompt_Engineering_on_Memecap_Templatic_Memes}, the GPT-4o prompt without template context achieved a BLEURT score of 0.525 on the whole test set. For comparison, a human-level model in \citep{bhavya2022analogy} scored 0.448 on analogy generation, suggesting that our result represents a human-level annotation. The method also attained a BERTscore of 0.879, indicating high-quality annotation. However, on other metrics (ChrF, ROUGE-L, and BLEU-4), our methods showed lower scores compared to the original paper. We suspect this is due to differences in sentence length and style between the generated captions and human annotations. To investigate further, we conducted additional experiments using templatic memes from MemeCap's test set. 
For few-shot prompting, we selected three meme examples that represent most of the dataset's meme types. As seen in Table \ref{tab:Prompt_Engineering_on_Memecap_Templatic_Memes}, while few-shot prompting helped standardize sentence length and style, scores for n-gram-based metrics remained significantly lower than those in the MemeCap paper. The best performance came from the 5-feature prompt, despite style differences. We believe this drop in performance for few-shot learning is due to the concise nature of the meme captions from MemeCap, which limits expressive space. Additionally, differences in lexical resources between GPT-4o and human annotators led to high semantic but low n-gram similarity. Refer to Appendix \ref{sec:Prompt_Engineering_on_Memecap_Templatic_Memes} for examples of the prompts.
\subsubsection{Conclusion}
In conclusion, while GPT-4o is on par with humans in image captioning, meme captioning, and embedded text extraction, it showed a strong bias in understanding figurative language in memes. Based on our findings in optimizing literary device labeling, we chose to use the \textbf{Three-Step-Reasoning prompt} with 12 
literary device labels for literary device labeling, and the \textbf{Zero-shot prompt} 
 (same as the prompt used for the human evaluation survey) with template context for captioning tasks and extracting embedded text.

\begin{table}[t]
  \centering
  \begin{tabular}{llllll}
    \hline
    \textbf{FT} & \textbf{Text}  & \textbf{R@1} & \textbf{R@5} & \textbf{R@10} & \textbf{{ Mean}}\\
    \hline
    \multirow{4}{*}{\makecell{$\times$}} &me. cap.   & 0.680 & 0.834 &0.874&0.796\\
    &im. cap.  & \textbf{0.419} & 0.592 &0.667&0.559\\
    &em. txt & {0.877} & 0.956 &0.965&0.933\\
    &title       & \textbf{0.222} & {0.352} &0.410&0.328\\
    \multirow{4}{*}{\makecell{$\checkmark$}} &me. cap.   & \textbf{0.770} & \textbf{0.892} &\textbf{0.919} &\textbf{0.860}\\
    &im. cap.  & 0.415 & \textbf{0.596} &\textbf{0.668}& \textbf{0.560}\\
    &em. txt & \textbf{0.946} & \textbf{0.978} &\textbf{0.982} &\textbf{0.969}\\
    &title       & 0.206 & \textbf{0.357} &\textbf{0.430} &\textbf{0.331}\\
    
    \hline

  \end{tabular}
  \caption{meme-text retrieval results for MemeCap between the original CLIP-ViT-L/14@336px model and its fine-tuned counterpart, denoted FT. Here, R@K Avg.~refers to the average R@K values for both meme-to-text (meme2text) and text-to-meme (text2meme) retrievals. Mean refers to Overall Mean, which represents the average score of R@1, R@5 and R@10. The labels me.~cap. and im.~cap. refer to meme caption and image caption, respectively, while em.~text denotes embedded text. The title column includes the meme title, for the MemeCap dataset it is specifically the post title.}
  \label{tab:evaluation_on_meme_text_tasks_memecap}
      \vspace{-3mm}
\end{table}

\section{Meme-Text Retrieval Fine-Tuning}

\begin{table}
  \centering
  \begin{tabular}{lllllll}
    \hline
    \textbf{FT} & \textbf{Text}  & \textbf{R@1} & \textbf{R@5} & \textbf{R@10} & \textbf{{Mean}}\\
    \hline
     \multirow{4}{*}{\makecell{$\times$}} &me. cap.   & {0.696} & 0.851 &0.890& 0.812\\
    &im. cap.  & 0.069 & {0.174} &0.283&0.175\\
    &em. txt & {0.882} & 0.949 &0.962&0.931\\
    &title       & 0.181 & {0.313} &0.390&0.295\\
    \multirow{4}{*}{\makecell{$\checkmark$}} &me. cap.   & \textbf{0.861} & \textbf{0.947} &\textbf{0.967} &\textbf{0.925}\\
    &im. cap.  & \textbf{0.073} & \textbf{0.197} &\textbf{0.329} &\textbf{0.200}\\
    &em. txt & \textbf{0.963} & \textbf{0.981} &\textbf{0.987} &\textbf{0.977}\\
    &title       & \textbf{0.210} & \textbf{0.338} &\textbf{0.415} &\textbf{0.321}\\
    \hline
  \end{tabular}
  \caption{meme-text retrieval results for CM50 between the original CLIP-ViT-L/14@336px model and its fine-tuned counterpart, denoted FT. The naming description can be found in Table \ref{tab:evaluation_on_meme_text_tasks_memecap}}
  \label{tab:evaluation_on_meme_text_tasks_cm50}
 \vspace{-3mm}
\end{table}

\subsection{Environment Setup}
As observed in the pilot study, VLM models face challenges in meme-text retrieval tasks. To address this, we fine-tuned CLIP-ViT-L/14@336px on our dataset and tested the model on the MemeCap test set. Only meme captions were used as the text. We followed the fine-tuning procedure outlined in \citep{kim2024fine}, adapting it to meet our study’s specific requirements. Initial hyperparameters were partially derived from (\citealp{rasheed2023fine}; \citealp{wei2023improving}), and we conducted a hyperparameter search using Ray Tune to further optimize performance. We used gradient accumulation to enable large batch-size updates on a single GPU. The hyperparameter search was conducted on a single Nvidia A100 80GB GPU, taking three days to complete.

\subsection{Results}
The fine-tuned CLIP model achieved a maximum improvement of 6.8\% in meme2text retrieval and 11.2\% in text2meme retrieval in terms of Recall@1, as shown in Table \ref{tab:pilot_study}.
During hyperparameter tuning, we explored configurations for the learning-rate scheduler with warm-up, batch size, the AdamW optimizer, and the number of epochs. The optimal hyperparameter configuration included: a cosine annealing learning-rate scheduler with a 1-epoch warm-up, starting at 1e-6, peaking at 1e-5, and returning to 1e-6; batch sizes of 2400 or 2048; an AdamW optimizer with weight decay set to 0.1, betas at 0.9 and 0.98, and epsilon at 1e-8; and a total of 20 epochs. 
We observed that in most fine-tuning trials, the model’s performance stabilized after 5 epochs. Between 5 and 20 epochs.
Additionally, if the learning rate exceeded 1e-4, the model’s performance dropped significantly, resulting in a much lower test score.

Finally, we evaluated both the original and fine-tuned CLIP models on multiple meme-text retrieval tasks, incorporating four text types: meme captions, image captions, humorous titles, and embedded texts, as shown in Table \ref{tab:evaluation_on_meme_text_tasks_memecap} and Table \ref{tab:evaluation_on_meme_text_tasks_cm50}. Fine-tuning CLIP solely on our meme caption data improves the model's performance over the original CLIP on most tasks. While the fine-tuned model performs slightly worse on image captions and meme titles, it achieves scores comparable to the baseline model. 
However, both models face challenges with meme titles in both datasets and image captions in the CM50. The difficulty with meme titles lies in their brevity and limited context, often consisting of just a few words, such as ``He did it.'' For the image captions in CM50, the captions primarily focus on describing the visual content while disregarding embedded text. As a result, these captions serve as straightforward descriptions of template images. Additionally, as shown in Figure \ref{fig:data_imbalance}, all the templates have over 50 instances, which can cause the target meme to be ranked beyond the 50th position, resulting in a low retrieval Recall@K score.

\section{Conclusion and Discussion}
In conclusion, our work contributes significantly to the field of meme understanding through three primary achievements: the creation of a novel dataset, CM50, for meme template annotation; the development of an automated annotation method closely aligned with human preferences; and the refinement of a retrieval model specifically for meme-text contexts. 
Our findings emphasize that our annotation methodology, leveraging GPT-4o with tailored prompts and template context, can achieve close to human-level performance in captioning tasks, underscoring its alignment with human preferences. Additionally, the use of template contexts has been demonstrated to enhance the model's accuracy and informativeness in labeling literary devices, which is essential for capturing the nuanced figurative language often present in memes. 

However, despite these advancements, GPT-4o still shows limitations in fully understanding complex literary devices, indicating an ongoing need for improvement in automated figurative language interpretation. Future research could explore domain adaptation techniques to overcome these challenges, aiming for more comprehensive meme comprehension.
Additionally, using the ``context+meme'' annotation method, future research can prioritize expanding datasets and standardizing annotation styles. With a sufficiently large dataset, combined with existing resources, researchers could pre-train an even more robust meme embedder, enhancing performance across numerous downstream tasks.

\section*{Acknowledgments}
This work was supported in part by the Pioneer Centre for AI, DNRF grant number P1.
\bibliography{custom}

\appendix

\section{Prompts For Model Selection}
\label{sec:Prompts_For_Model_Selection}
Table \ref{tab:Prompts_For_Model_Selection} lists the prompts used for meme annotations in human preference evaluations.
\begin{table*}
  \centering
  \begin{tabularx}{\textwidth}{|m{2cm}|X|}
    \hline
    \makecell{\textbf{Model}} & \makecell{\textbf{Prompt}}\\
    \hline
\makecell{GPT-4o} & Here is the context of the meme: \{\}. First, based the given context, read the text in this image and explain the meme. Then, provide information for the following categories: 
    
 Visual Elaboration (focus on the main content):
 
 Detected Text: 
 
 Meaning of the Meme (briefly): 
 
 Then, choose the most suitable literary device from the given category words: sarcasm, allegory, alliteration, allusion, amplification, anagram, analogy, anthropomorphism, antithesis, chiasmus, circumlocution, euphemism, hyperbole, imagery, metaphor, onomatopoeia, oxymoron, paradox, personification, portmanteau, pun, satire, simile, and symbolism. If no suitable word, use ``None'' as the category word. Only reply with the chosen word. 
 Finally, choose the most suitable emotion word from the given category words: fear, anger, joy, sadness, surprise, disgust, guilt, contempt, shame, embarrassment, envy, jealousy, love, hate, and interest. If no suitable word, use ``None'' as the category word. Only reply with the chosen word.\\
    \hline
    \makecell{LlaVA-1.6} & Here is the context of the meme: \{\}. First, based the given context, read the text in this image and explain the meme. Then, provide information for the following categories:
    
 Visual Elaboration (focus on the main content): 
 
 Detected Text: 
 
 Meaning of the Meme (briefly):
 
 Literary Device (category words only):

Emotion (category words only):\\
    \hline
  \end{tabularx}
  \caption{\label{tab:Prompts_For_Model_Selection}
    In this context, ``Visual Elaboration'' refers to ``Image caption,'' ``Detected Text'' to ``Embedded text,'' and ``Meaning of the meme'' to ``Meme caption.'' As LlaVA-1.6 struggles with long prompts in instruction following, we removed candidate labels for literary devices and emotions.  
  }
\end{table*}

\section{Human Preference Evaluation}
\label{sec:Human_Preference_Evaluation}
As shown in Table \ref{tab:Human_Preference_Evaluation}, the survey includes five questions per meme, each structured with an ``Explanation of the question and criteria for answering.'' For example:\\
\textbf{Meme caption}:
\begin{itemize}
    \item [a)] A meme caption explains the humor of a meme. 
\end{itemize}
\textbf{Criteria}:
\begin{itemize}
    \item [a)] Accuracy: Does the caption convey the humor correctly?
    \item [b)] Relevance: Is the caption fully related to the humor?
\end{itemize}
\begin{table*}
  \centering
  \begin{tabular}{ll|ll|ll|ll}
    \hline
    \textbf{memes} & \textbf{subtasks} & \multicolumn{2}{|c|}{\textbf{GPT-4o}}& \multicolumn{2}{|c|}{\textbf{LlaVA v1.6 34B}}& \multicolumn{2}{c}{\textbf{LlaVa v1.6 7B 32bit}} \\
    \hline
    &&With &Without&With &Without&With &Without\\
    \hline
    \multirow{5}{*}{\includegraphics[width = 2.4cm]{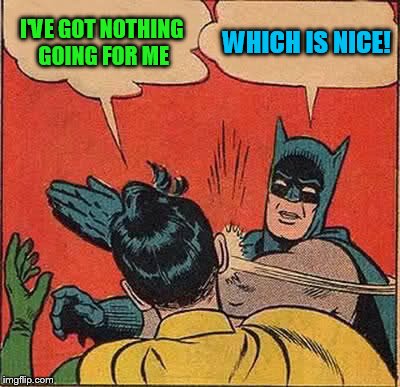}} & embedded text & 4 & 3 & 3 & 3 & 0 & 2 \\
    & image caption & 1 & 1 & 2 & 0 & 0 & 0 \\
    & meme caption & 3 & 0 & 1 & 0 & 0 & 0 \\
    & literary devices & 3 & 0 & 3 & 3 & 1 & 0 \\
    & emotions & 1 & 0 & 1 & 1 & 1 & 3 \\
    \hline
    \multirow{5}{*}{\includegraphics[width = 2.4cm]{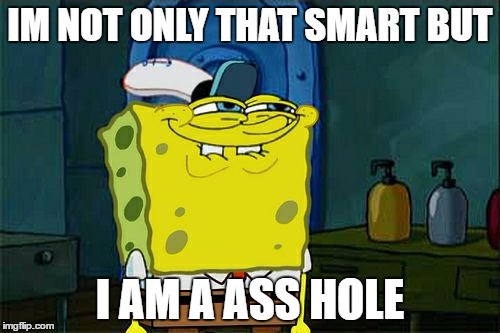}} & embedded text & 4 & 3 & 3 & 0 & 1 & 2 \\
    & image caption & 1 & 1 & 1 & 1 & 0 & 0 \\
    & meme caption & 0 & 3 & 0 & 1 & 0 & 0 \\
    & literary devices & 3 & 2 & 2 & 2 & 0 & 1 \\
    & emotions & 0 & 0 & 3 & 3 & 0 & 0 \\
    \hline
    \multirow{5}{*}{\includegraphics[width = 2.4cm]{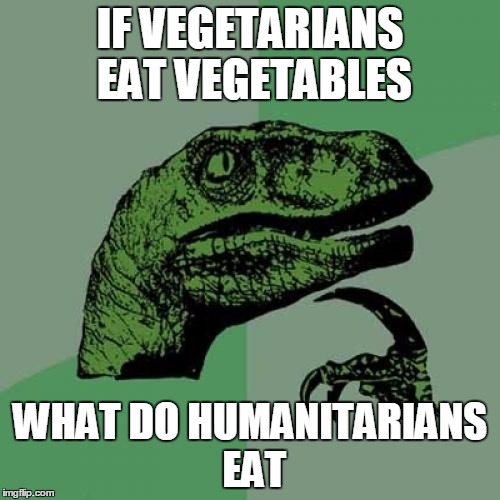}} & embedded text & 4 & 1 & 3 & 0 & 1 & 1 \\
    & image caption & 0 & 0 & 1 & 0 & 0 & 2 \\
    & meme caption & 1 & 3 & 0 & 0 & 0 & 0 \\
    & literary devices & 2 & 2 & 2 & 1 & 2 & 1 \\
    & emotions & 0 & 0 & 1 & 1 & 1 & 3 \\
    \hline
    \multirow{5}{*}{\includegraphics[width = 2.4cm]{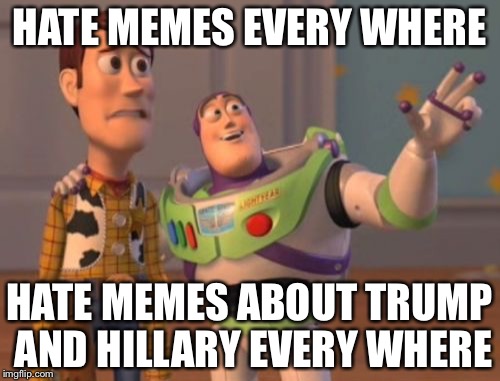}} & embedded text & 4 & 3 & 3 & 0 & 2 & 0 \\
    & image caption & 2 & 0 & 2 & 0 & 0 & 0 \\
    & meme caption & 1 & 1 & 0 & 2 & 0 & 0 \\
    & literary devices & 2 & 2 & 1 & 1 & 1 & 1 \\
    & emotions & 1 & 3 & 3 & 0 & 0 & 0 \\
    \hline
    \multirow{5}{*}{\includegraphics[width = 1.8cm]{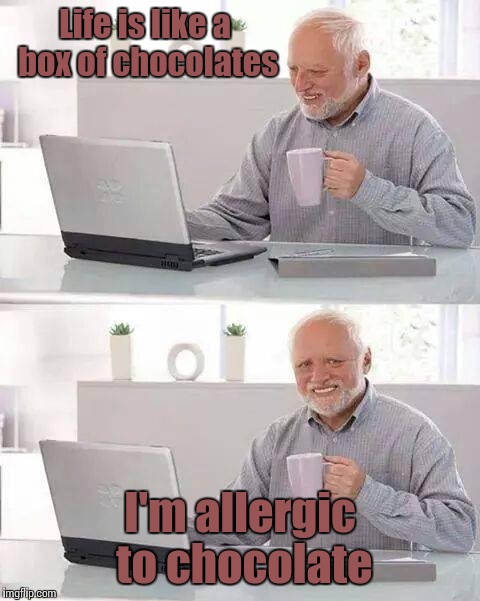}} & embedded text & 4 & 2 & 3 & 3 & 1 & 0 \\
    & image caption & 2 & 2 & 0 & 0 & 0 & 0 \\
    & meme caption & 3 & 1 & 0 & 0 & 0 & 0 \\
    & literary devices & 1 & 1 & 3 & 3 & 2 & 4 \\
    & emotions & 2 & 0 & 3 & 3 & 1 & 1 \\
    \hline
    \textbf{Total} & & \textbf{49} & \textbf{34} & \textbf{44} & \textbf{28} & \textbf{14} & \textbf{21} \\
    \hline
  \end{tabular}
  \caption{\label{tab:Human_Preference_Evaluation}
    This table presents the results of a human preference evaluation survey, comparing three models with two prompt types: one including template context (the ``With'' condition) and one excluding it (the ``Without'' condition). Survey questions, except for image caption and meme caption, are multi-option questions.
  }
\end{table*}

\newpage
\section{Prompt Engineering on Figmemes Test Set}
\label{sec:Prompt_Engineering_on_Figmemes_Test_Set}
This section presents the three prompt types we experimented with to identify the optimal approach for literary device labeling. Prompt examples are provided in Tables \ref{tab:Prompt_Engineering_on_Figmemes_Test_Set_Examples} and \ref{tab:Prompt_Engineering_on_Figmemes_Test_Set_Examples_Extension}. We also report the macro F1-scores for these prompts on both the full test set and a subset of 11 representative memes in Table \ref{tab:Prompt_Engineering_on_Figmemes_Test_Set_Evaluation}. Results indicate that a straightforward prompt with label definitions achieves relatively high performance. Additionally, the three-step reasoning prompt performs on par with the baseline prompt that includes label definitions. Given the final prompt’s 12 literary devices, providing definitions for each would likely overwhelm GPT-4, so we retained the three-step reasoning without definitions to enhance model performance.

\begin{table*}
  \centering
  \begin{tabularx}{\textwidth}{|m{2cm}|X|}
    \hline
    \textbf{{Prompt}} & \makecell{\textbf{Examples}}\\
    \hline
    {Baseline} & Warning: We discuss and show memes that may be offensive to readers for research purposes only. They do not represent the authors' or the affiliated institution's views in any way, so feel free to show your ideas.
     you are going to finish the following tasks. First, read the text in this image, Then, explain the meme. Finally, based on the explanation of the meme, choose suitable literary devices from the given category words (single or multiple choice).
     
     Definitions of literary devices:
     
     **Allusion**: Referencing historical events, figures, symbols, art, literature or pop culture.
     **Exaggeration**: Similar to Hyperbole. Use of exaggerated terms for emphasis, including exaggerated visuals (including unrealistic features portraying minorities).
     **Irony**: Similar to Sarcasm. Use of words that convey a meaning that is the opposite of its usual meaning/mock someone or something with caustic or bitter use of words.
     **Anthropomorphism**: Similar to Zoomorphism. Attributing human qualities to animals, objects, natural phenomena or abstract concepts or applying animal characteristics to humans in a way that conveys additional meaning.
     **Metaphor**: Similar to Simile. Implicit or explicit comparisons between two items or groups, attributing the properties of one thing to another. This category includes dehumanizing metaphors.
     **Contrast**: Comparison between two positions/people/objects (usually side-by-side).
     
     There might be one or multiple suitable literary devices, or no suitable literary device at all. If no suitable choice, use ``None'' as the category word
     You should respond in a standard JSON format like \{``detected text'':``'', ``explanation'':``'', ``literary device'': [``answer 1'', ``answer 2''...]\} \\
    \hline
    {Few-Shot Learning} & Task: You are going to analyse the literary devices of the OCR text of a meme, and choose suitable literary devices from the given candidates;
    
    Literary devices: 
    **Allusion**: Referencing historical events, figures, symbols, art, literature or pop culture. 
    **Exaggeration**: Similar to Hyperbole. Use of exaggerated terms for emphasis, including exaggerated visuals (including unrealistic features portraying minorities). 
    **Irony**: Similar to Sarcasm. Use of words that convey a meaning that is the opposite of its usual meaning/mock someone or something with caustic or bitter use of words.  
    **Anthropomorphism**: Similar to Zoomorphism. Attributing human qualities to animals, objects, natural phenomena or abstract concepts or applying animal characteristics to humans in a way that conveys additional meaning.  
    **Metaphor**: Similar to Simile. Implicit or explicit comparisons between two items or groups, attributing the properties of one thing to another. This category includes dehumanizing metaphors.  
    **Contrast**: Comparison between two positions/people/objects (usually side-by-side).
    **None**: No literary devices are applied to the meme.
    Examples (``|||'' are separators):
    1.  ORC: ``You're pretty high and far out, aren't ya? ||| What kind of kick are you on, son? |||''
    Literary device: [``none'']
    2. OCR: ``US Marine Training Meanwhile in Russia ||| MemeCenter te ||| VIA FUNPICS.ME |||''
    Literary device: [``contrast'']
    3. OCR: ``6 year old: I love reading fantasy books ||| He: (trying to impress her) writes Quran ||| dco |||''
    Literary device: [``allusion'', ``irony'', ``metaphor'']
    
    Analyse the following OCR: ``''
    
    Follow the JSON format: 
    \{
    ``literary device'': [``word 1'', …],
    \} \\
    \hline
  \end{tabularx}
  \caption{\label{tab:Prompt_Engineering_on_Figmemes_Test_Set_Examples}
    The Baseline prompt focuses solely on literary device labeling, excluding elements of other labeling tasks. The Three-Step-Reasoning prompt, however, includes multi-label choice, answer extraction, and choice-by-choice comparison. This prompt enables the model to approach the task from two perspectives: inferring labels based on the meme content and verifying the meme’s validity against the labels. Labels that are validated from both directions are selected as the final labels for the meme.      
  }
\end{table*}

\begin{table*}
  \centering
  \begin{tabularx}{\textwidth}{|m{2cm}|X|}
    \hline
    \textbf{{Prompt}} & \makecell{\textbf{Examples}}\\
    \hline
    {Three-Step-Reasoning prompt} & ``You are a masterful assistant in the interpretation of online memes, their style of literary devices, their meaning and humour..''

    <The list of choices and the definitions of literary devices>: 
    **Allusion**: Referencing historical events, figures, symbols, art, literature or pop culture.
    **Exaggeration**: Similar to Hyperbole. Use of exaggerated terms for emphasis, including exaggerated visuals (including unrealistic features portraying minorities).
    **Irony**: Similar to Sarcasm. Use of words that convey a meaning that is the opposite of its usual meaning/mock someone or something with caustic or bitter use of words.
    **Anthropomorphism**: Similar to Zoomorphism. Attributing human qualities to animals, objects, natural phenomena or abstract concepts or applying animal characteristics to humans in a way that conveys additional meaning. 
    **Metaphor**: Similar to Simile. Implicit or explicit comparisons between two items or groups, attributing the properties of one thing to another. This category includes dehumanizing metaphors. 
    **Contrast**: Comparison between two positions/people/objects (usually side-by-side). 
    
    ``<Multiple Choice> Please select one or multiple labels from the above list that are applied to the meme:''
    Your answer:
    
    ``<Extraction of answer> Extract the suitable labels for the input meme and the multiple choice question above:''
    Your answer:
    
    ``<Choice by choice comparison> Compare each label with the meme and decide if this label could explain the meme:''
    Your answer:
    
    ``Finally output your answer in the format:''
    \{
    ``literary device'':[``allusion'', ...]
    \} \\
    \hline
  \end{tabularx}
  \caption{\label{tab:Prompt_Engineering_on_Figmemes_Test_Set_Examples_Extension}
    The extension of Table \ref{tab:Prompt_Engineering_on_Figmemes_Test_Set_Examples}.      
  }
\end{table*}

\begin{table*}
  \centering
  \begin{tabularx}{\textwidth}{m{3cm}Xm{2.5cm}}
    \hline
    \textbf{Prompt Type} & \textbf{Details} & \textbf{macro F1-score} \\
    \hline
    \multirow{3}{*}{\makecell{Baseline}} & With literary device definitions on the whole test set & \textbf{0.39} \\
    & With literary device definitions & \textbf{0.40}  \\
    & Without literary device definitions & 0.31  \\
    \hline
    \multirow{6}{*}{\makecell{Few-Shot Learning}} & With meme explanation examples &  0.27\\
    & With explanation and literary device pairs examples & 0.28  \\
    & With meme image and literary device pairs examples &  0.31  \\
    & With explanation examples for each literary device examples & 0.25 \\
    & With OCR and literary device pairs examples & 0.35 \\
    & With OCR text examples for each literary device examples & 0.36 \\
    \hline
    \multirow{2}{*}{\makecell{Three-Step- \\Reasoning prompt}} & base & \textbf{0.39}\\
    & With a critical personality  & 0.34\\
    \hline
  \end{tabularx}
  \caption{\label{tab:Prompt_Engineering_on_Figmemes_Test_Set_Evaluation}
    This table presents the macro F1-scores of various prompts tested for literary device labeling. Both the few-shot learning and Three-Step-Reasoning prompts build on the baseline prompt. The two baseline prompts are identical but applied to different data scales: the first on the entire test set and the second on a subset of 11 representative samples. The similarity in macro F1 scores between these two baseline prompts suggests that the selected samples effectively approximate the distribution of the full test set.   
  }
\end{table*}
\newpage
\section{Prompt Engineering on MemeCap Templatic Memes}
\label{sec:Prompt_Engineering_on_Memecap_Templatic_Memes}
This section presents our prompt engineering results on the Memecap test set. We observed that caption length and description style impact scores on n-gram-based metrics, prompting us to further evaluate the performance of few-shot prompts. \\
The prompts used in these experiments build on those from the model selection phase. We tested the effects of few-shot versus zero-shot prompts, with and without template context. The zero-shot prompts remained the same as that in model selection phase, while the text-only few-shot prompts focused exclusively on meme caption-related tasks, omitting other tasks to help GPT-4 better concentrate on meme captioning—specifically, explaining meme meanings. Results are shown in Table \ref{tab:Prompt_Engineering_on_Memecap_Templatic_Memes}

\begin{table*}
  \centering
  \begin{tabular}{lllllll}
    \hline
    \textbf{Prompt}&\textbf{Metrics}&\textbf{BLEURT} &\textbf{BERTscore}&\textbf{ChrF} &\textbf{ROUGE-L}&\textbf{BLEU-4} \\
    \hline
    \multirow{5}{2.4cm}{Text-only few-shot with the template context} & Extended predictions & 0.313&0.862&22.631&0.142&0  \\
    & Concatenated references & 0.369&0.858&15.063&0.118&0  \\
    & Multi-reference & &0.873 && 0.191 &0  \\
    & Best match & 0.352 &  &  &  &   \\
    & Fully-concatenated & 0.467 &0.846& 28.032& 0.158&\textbf{0.021} \\
    \hline
    \multirow{5}{2.4cm}{Text-only few-shot without the template context} & Extended predictions & 0.309&0.864&21.821&0.138&0  \\
    & Concatenated references & 0.364& 0.859& 14.159& 0.115&0  \\
    & Multi-reference & &0.873&&0.190 &0  \\
    & Best match & 0.344 &  &  &  &   \\
    & Fully-concatenated & 0.449& 0.847& 27.225& 0.153& 0.002  \\
    \hline
    \multirow{5}{2.4cm}{Zero-shot with the template context} & Extended predictions & 0.254& 0.853& 20.049& 0.087 &0  \\
    & Concatenated references & 0.332& 0.842& 16.220& 0.090 &0  \\
    & Multi-reference &  & 0.862 &  & 0.126 & 0  \\
    & Best match & 0.286 &  &  &  &   \\
    & Fully-concatenated & 0.391& 0.836& 24.414& 0.113& 0  \\
    \hline
    \multirow{5}{2.4cm}{Zero-shot without the template context} & Extended predictions & 0.256& 0.851& 19.498& 0.087 &0  \\
    & Concatenated references & 0.331& 0.842& 15.101& 0.088 &0 \\
    & Multi-reference &  & 0.859 &  & 0.122 & 0  \\
    & Best match & 0.290 &  &  &  &   \\
    & Fully-concatenated & 0.396& 0.836& 24.207& 0.104 & 0  \\
    \hline
    \multirow{5}{2.4cm}{Zero-shot with 5 tasks on the whole test set} & Extended predictions & 0.471& 0.869& 27.860& 0.173& 0.002  \\
    & Concatenated references & 0.474& 0.864& 0.173& 0.005  \\
    & Multi-reference &  & 0.879 &  & \textbf{0.229} &   \\
    & Best match & 0.525& 0.879& 23.779 &   \\
    & Fully-concatenated &  &  &  &  &   \\
    \hline
    \multirow{5}{2.4cm}{Zero-shot on 5 tasks with the template context} & Extended predictions & 0.472& 0.870& 28.653& 0.178& 0.002  \\
    & Concatenated references & 0.479& 0.863& 23.460& 0.180& 0.006 \\
    & Multi-reference &  & 0.879 &  & 0.228 &   \\
    & Best match & 0.523& 0.879 &  &  &   \\
    & Fully-concatenated &  &  &  &  &   \\
    \hline
    \multirow{5}{2.4cm}{Zero-shot on 5 tasks without the template context} & Extended predictions & 0.487& 0.872& \textbf{29.083}& 0.179& 0.002  \\
    & Concatenated references & 0.489& 0.866& 24.408& 0.182& 0.006  \\
    & Multi-reference &  & \textbf{0.881} &  & 0.231 & 0.007  \\
    & Best match & \textbf{0.540}& \textbf{0.881} &  &  &   \\
    & Fully-concatenated &  &  &  &  &   \\
    \hline
  \end{tabular}
  \caption{\label{tab:Prompt_Engineering_on_Memecap_Templatic_Memes}
    This table shows the results of prompt engineering on the templatic memes in the test set of MemeCap. Extended predictions: For single-prediction vs. multi-reference cases, predictions are duplicated to match the number of references; Concatenated references: For single-prediction vs. multi-reference cases, all references are concatenated into a single caption; Multi-reference: For single-prediction vs. multi-reference cases, references are used as-is, applying only metrics that support multiple references; Best match: For single-prediction vs. multi-reference cases, the reference with the best match is selected for final score calculation; Fully-concatenated: For multi-prediction vs. multi-reference cases, all predictions and references are concatenated into single entries separately, with metrics calculated on the resulting text. 
  }
\end{table*}

\newpage
\section{Prompt Engineering on Figmemes Templatic Memes}
\label{sec:Prompt_Engineering_on_Figmemes_Templatic_Memes}
This section presents the results of applying the three-step reasoning prompt on templatic memes in the Figmemes test set. We initially assessed the informativeness of the template context, then modified the labels to align with our target label set, removing label definitions in the process. To compute the macro F1 score, we mapped our labels to the six Figmemes labels, as outlined in Table \ref{tab:Prompt_Engineering_on_Figmemes_Templatic_Memes_Map}.\\
Our findings indicate that the template context is informative and helps improve accuracy. However, the extended prompt did not achieve a high macro F1 score; These scores are provided as a reference for future research. All scores are detailed in Table \ref{tab:Prompt_Engineering_on_Figmemes_Templatic_Memes}.

\begin{table*}
  \centering
  \begin{tabularx}{\textwidth}{m{3cm}Xm{2.5cm}}
    \hline
    \textbf{Prompt Type} & \textbf{Details} & \textbf{macro F1-score} \\
    \hline
    \multirow{4}{*}{\makecell{Three-Step- \\Reasoning prompt}} & With the template context and the label definitions  & \textbf{0.33}\\
    & Without the template context but with the label definitions   & 0.32\\
    & With the template context and 12 literary devices   & 0.25\\
    & With the template context and 12 literary devices and with the emphasis on avoiding generating ``sarcasm'' and ``irony''   & 0.26\\
    \hline
  \end{tabularx}
  \caption{\label{tab:Prompt_Engineering_on_Figmemes_Templatic_Memes}
    This table presents the macro F1-scores of prompts applied to templatic memes in the Figmemes test set. All prompts are based on the Three-Step-Reasoning prompt, tested under four different conditions.   
  }
\end{table*}

\begin{table*}
  \centering
  \begin{tabularx}{\textwidth}{Xm{2.5cm}}
    \hline
    \textbf{Source Labels} & \textbf{Target Labels} \\
    \hline
    irony, sarcasm& irony\\
    \hline
    anthropomorphism, personification & anthrop\\
    \hline
    contrast, paradox, antithesis, oxymoron & contrast\\
    \hline
    metaphor, simile & metaphor\\
    \hline
    exaggeration, amplification & exaggeration\\
    \hline
    Allusion & allusion\\
    \hline
    anagram, pun, allegory, alliteration, analogy, antithesis, chiasmus, circumlocution, euphemism, imagery, onomatopoeia, portmanteau,  symbolism, satire & None\\
    \hline
  \end{tabularx}
  \caption{\label{tab:Prompt_Engineering_on_Figmemes_Templatic_Memes_Map}
    This table outlines the mapping of our labels to the Figmemes labels. The label ``none'' indicates that the corresponding source labels were disregarded in the mapping process.  
  }
\end{table*}

\newpage
\section{CM50 Dataset details}
\label{sec:CM50_Dataset_details}
\begin{figure*}[t]
  \includegraphics[width=0.48\linewidth]{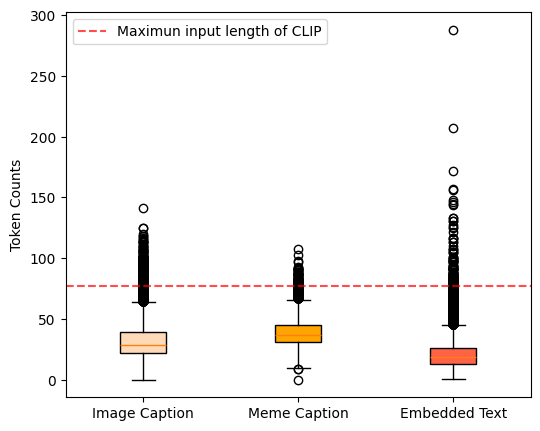} \hfill
  \includegraphics[width=0.48\linewidth]{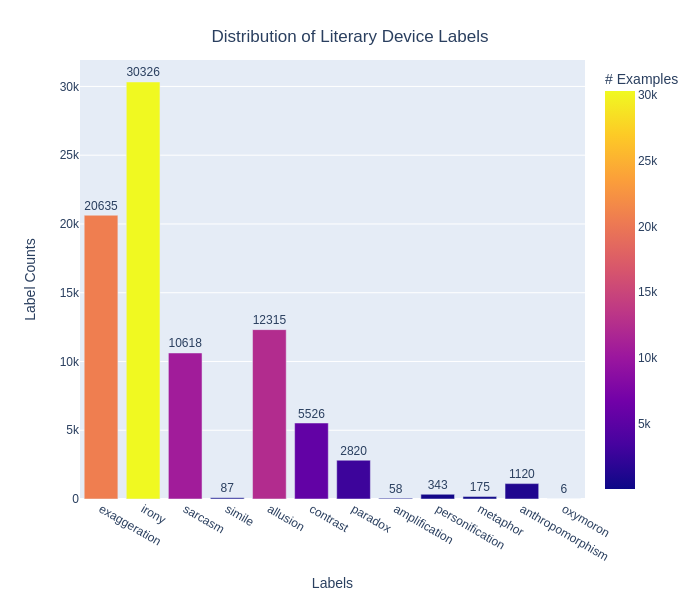}
  \caption {Left: Distribution of lengths (in tokens) for image captions, meme captions, and embedded text;
  Right: Statistical distribution of literary device labels across the dataset.}
  \label{fig:annotation_statistics}
\end{figure*}

\begin{figure*}[t]%
    \centering
    \includegraphics[width=\linewidth]{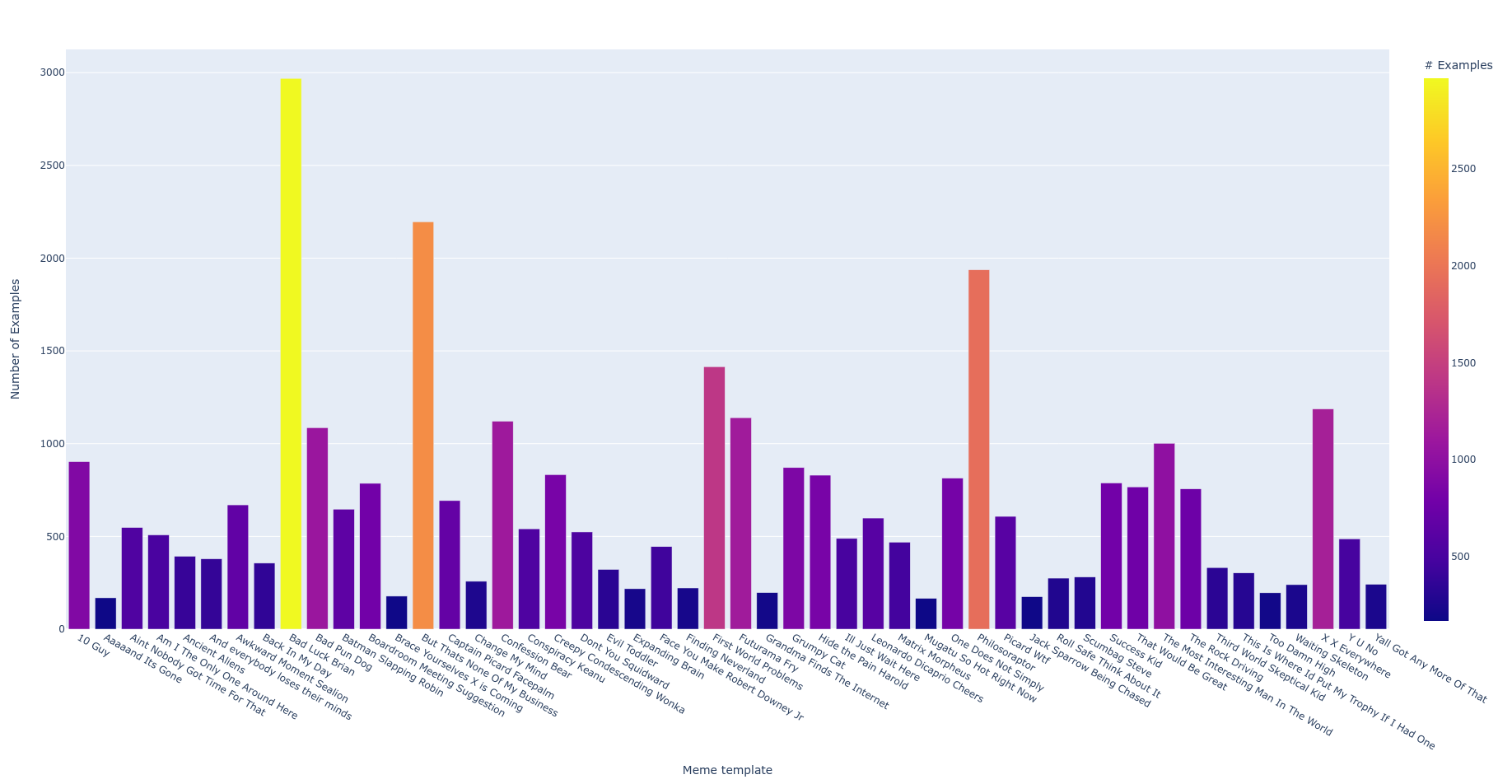}
    \caption{CM50 dataset statistics of different templates present in our dataset. }
    \label{fig:data_imbalance}
\end{figure*}

Here, we present the names of the 50 templates, as shown in Table \ref{tab:meme_templates_cm50}, and their instance counts, as shown in Figure \ref{fig:data_imbalance}.
We also present different dataset statistics for the 33k labelled memes with both their meme captions, image captions and OCR text, as well as their designated Literary Devices labels in \ref{fig:annotation_statistics}.

\begin{table*}
  \centering
  \begin{tabularx}{\textwidth}{Xm{2.5cm}}
    \hline
    \makecell{\textbf{Meme Template}} \\
    \hline
    10-Guy, Aaaaand-Its-Gone, Aint-Nobody-Got-Time-For-That, Am-I-The-Only-One-Around-Here \\
    \hline
    Ancient-Aliens, And-everybody-loses-their-minds, Awkward-Moment-Sealion, Back-In-My-Day\\
    \hline
    Bad-Luck-Brian, Bad-Pun-Dog, Batman-Slapping-Robin, Boardroom-Meeting-Suggestion\\
    \hline
    Brace-Yourselves-X-is-Coming, But-Thats-None-Of-My-Business, Captain-Picard-Facepalm\\
    \hline
    Change-My-Mind, Confession-Bear, Conspiracy-Keanu, Creepy-Condescending-Wonka\\
    \hline
    Dont-You-Squidward, Evil-Toddler, Expanding-Brain, Face-You-Make-Robert-Downey-Jr\\
    \hline
    Finding-Neverland, First-World-Problems, Futurama-Fry, Grandma-Finds-The-Internet, Grumpy-Cat\\
    \hline
     Hide-the-Pain-Harold, Ill-Just-Wait-Here, Leonardo-Dicaprio-Cheers, Matrix-Morpheus\\
    \hline
    Mugatu-So-Hot-Right-Now, One-Does-Not-Simply, Philosoraptor, Picard-Wtf\\
    \hline
    Jack-Sparrow-Being-Chased, Roll-Safe-Think-About-It, Scumbag-Steve, Success-Kid\\
    \hline
    That-Would-Be-Great, The-Most-Interesting-Man-In-The-World, The-Rock-Driving\\
    \hline
    Third-World-Skeptical-Kid, This-Is-Where-Id-Put-My-Trophy-If-I-Had-One, Too-Damn-High \\
    \hline
    Waiting-Skeleton, X-X-Everywhere, Y-U-No, Yall-Got-Any-More-Of-That \\
    \hline
  \end{tabularx}
  \caption{\label{tab:meme_templates_cm50}
    The meme templates found in CM50
  }
\end{table*}

\end{document}